%% file: acl_latex.tex
\pdfoutput=1

\documentclass[11pt]{article}

\usepackage[]{acl2023}

\usepackage{times}
\usepackage{latexsym}
\usepackage{pgfplots}
\usepackage{lipsum} 
\usepackage{multirow}
\usepackage{colortbl}
\usepackage{booktabs}
\usepackage{amssymb}
\usepackage{diagbox}
\usepackage[normalem]{ulem}
\usepackage{makecell}

\usepackage{tikz}

\usepackage[draft]{todonotes}

\usepackage{array,booktabs,arydshln,xcolor}
\usepackage{verbatim}
\usepackage{url}
\usepackage{graphicx}

\definecolor{myred}{RGB}{180, 31, 39}
\definecolor{light-gray}{gray}{0.8}

\definecolor{lightyellow}{HTML}{eb9634}
\definecolor{lightred}{HTML}{FFCCCC}
\definecolor{lightpurple}{HTML}{CCCCFF}
\definecolor{lightblue}{rgb}{.90,.92,1}
\definecolor{lichen}{rgb}{.91,.95,0.83}

\usepackage{graphicx}
\usepackage{float}
\definecolor{hred}{RGB}{247,202,197}
\definecolor{horange}{RGB}{242,217,198}
\definecolor{hyellow}{RGB}{247,225,147}
\definecolor{hlgreen}{RGB}{200,207,141}
\definecolor{hdgreen}{RGB}{172,215,156}
\definecolor{hlblue}{RGB}{181,228,227}
\definecolor{hdblue}{RGB}{188,220,246}
\definecolor{hpurple}{RGB}{219,218,246}

\definecolor{hdblue}{RGB}{188,220,236}
\definecolor{hdblue2}{RGB}{188,220,236}

\definecolor{mygray}{HTML}{C0C0C0}

\usepackage{stfloats}
\usepackage[T1]{fontenc}

\usepackage[utf8]{inputenc}

\usepackage{microtype}

\title{\textsc{MSciNLI}: A Diverse Benchmark for Scientific Natural Language Inference}

\author{Mobashir Sadat \mbox{   }\mbox{   }\mbox{   }\mbox{   } Cornelia Caragea\\
  Computer Science \\
  University of Illinois Chicago \\
  {\color{blue}\texttt{msadat3@uic.edu \mbox{    }  cornelia@uic.edu}} 
  }

\begin{document}
\maketitle
\begin{abstract}

\looseness=-1
The task of scientific Natural Language Inference (NLI) involves predicting the semantic relation between two sentences extracted from research articles. This task was recently proposed along with a new dataset called \textsc{SciNLI} derived from papers published in the computational linguistics domain. In this paper, we aim to introduce diversity in the scientific NLI task and present \textsc{MSciNLI}, a dataset containing $132,320$ sentence pairs extracted from five new scientific domains. 
The availability of multiple domains makes it possible to study domain shift for scientific NLI. We establish strong baselines on \textsc{MSciNLI} by fine-tuning Pre-trained Language Models (PLMs) and prompting Large Language Models (LLMs). The highest Macro F1 scores of PLM and LLM baselines are $77.21\%$ and $51.77\%$, 
respectively, illustrating that \textsc{MSciNLI} is challenging for both types of models. Furthermore, we show that domain shift degrades the performance of scientific NLI models which demonstrates the diverse characteristics of different domains in our dataset. Finally, we use both scientific NLI datasets in an intermediate task transfer learning setting and show that they can improve the performance of downstream tasks in the scientific domain. We make our dataset and code available on Github.\footnote{\url{https://github.com/msadat3/MSciNLI}}

\end{abstract}

\input{Camera_ready}

\end{document}

%% file: Camera_ready.tex
\section{Introduction}
\looseness=-1
Natural Language Inference (NLI) \cite{bowman-etal-2015-large} or Textual Entailment is the task of recognizing the semantic relation between a pair of sentences where the first sentence is called premise and the second sentence is called hypothesis. Traditional NLI datasets such as SNLI \cite{bowman-etal-2015-large}, MNLI \cite{williams-etal-2018-broad}, SICK \cite{marelli-etal-2014-sick}, and ANLI \cite{nie2019adversarial} classify the premise-hypothesis pairs into one of three classes indicating whether the hypothesis entails, contradicts or is neutral to the premise. These datasets have been used both as a benchmark for Natural Language Understanding (NLU) and to improve downstream tasks such as fact verification \cite{martin2022facter} and fake news detection \cite{sadeghi2022fake}. In addition, they have aided in the advancement of representation learning \cite{conneau2017supervised}, transfer learning \cite{pruksachatkun-etal-2020-intermediate}, and multi-task learning \cite{liu-etal-2019-multi-task}.

However, since the examples in these datasets are derived from non-specialized domains, e.g., image captions, they do not capture the unique linguistic characteristics of different specialized domains such as the scientific domain. More recently, \citet{sadat-caragea-2022-scinli} introduced the task of scientific NLI along with the first dataset for this task named \textsc{SciNLI}, which contains 
$107,412$ sentence pairs 
extracted exclusively from scientific papers related to 
computational linguistics published in the ACL anthology \cite{bird2008acl, radev-etal-2009-acl}. To capture the inferences that frequently occur in scientific text, \citet{sadat-caragea-2022-scinli} extended the three classes in traditional NLI 
to four classes for scientific NLI---\textsc{entailment}, \textsc{reasoning}, \textsc{contrasting}, and \textsc{neutral}. Since its introduction, \textsc{SciNLI} has gained great interest in the research community \cite{wang2022unsupervised, deka2022evidence, wu-etal-2023-characterizing}.

Despite introducing a challenging task and enabling the exploration of NLI with scientific text, \textsc{SciNLI} lacks the diversity to serve as a general purpose scientific NLI benchmark because it is limited to a single domain (ACL). Moreover, due to the unavailability of multiple domains,  \textsc{SciNLI} is not suitable for studying domain adaptation and transfer learning on scientific NLI. 

To this end, in this paper, we propose \textbf{\textsc{MSciNLI}}, a scientific NLI dataset containing $132,320$ sentence pairs extracted from papers published in five different domains: ``Hardware'', ``Networks'', ``Software \& its Engineering'', ``Security \& Privacy'', and ``NeurIPS.'' Similar to \citet{sadat-caragea-2022-scinli}, we use a distant supervision method that exploits the linking phrases between sentences in scientific papers to construct a large training set and directly use these potentially noisy sentence pairs during training. For the test and development sets, we manually annotate $4,000$ and $1,000$ examples, respectively, to create high quality evaluation data for scientific NLI.

We evaluate the difficulty of \textsc{MSciNLI} by experimenting with a \textsc{BiLSTM} based model. We then establish strong baselines on \textsc{MSciNLI} by 
a) fine-tuning four transformer based Pre-trained Language Models (PLMs): \textsc{BERT} \cite{devlin-etal-2019-bert}, \textsc{SciBERT} \cite{beltagy-etal-2019-scibert}, \textsc{RoBERTa} \cite{liu2019roberta} and \textsc{XLNet} \cite{yang2019xlnet}; and b) prompting two Large Language Models (LLMs) in both zero-shot and few-shot settings: \textsc{Llama-2} \cite{touvron2023llama}, and \textsc{Mistral} \cite{jiang2023mistral}. Furthermore, we provide a comprehensive investigation into the robustness of scientific NLI models by evaluating their performance under domain-shift at test time. Finally, we explore both \textsc{SciNLI} and \textsc{MSciNLI} in an intermediate task transfer learning setting \cite{pruksachatkun-etal-2020-intermediate} to evaluate their usefulness in improving the performance of other downstream tasks.

Our key findings are: a) \textsc{MSciNLI} is more challenging than \textsc{SciNLI}; b) the best performing PLM baseline, which is based on \textsc{RoBERTa}, shows a Macro F1 of $77.21\%$ on \textsc{MSciNLI} indicating the challenging nature of the task and a substantial headroom for improvement; c) the best performing LLM baseline with \textsc{Llama-2} shows a Macro F1 of only $51.77\%$ indicating that our dataset can be used to benchmark the NLU and complex reasoning capabilities of powerful LLMs; d) domain-shift at test time reduces the performance; and e) diversity in the scientific NLI datasets helps toimprove the performance of downstream tasks.

\section{Related Work}
\vspace{-2mm}

Since the introduction of the NLI task, many datasets derived from different data sources have been made available. Datasets such as RTE \cite{dagan06rte} and SICK \cite{marelli-etal-2014-sick} were instrumental in the progress of NLI research in its earlier days. However, the training set sizes of these datasets are too small for large scale deep learning modeling. \textsc{SNLI} \cite{bowman-etal-2015-large} was introduced as a large dataset for NLI. SNLI contains $570$K sentence pairs where premises are extracted from image captions and human crowdworkers were employed to write the hypotheses and assign the labels. While \textsc{SNLI} is significantly larger than all other prior datasets, due to the premises being extracted from a single source, it lacks the diversity to serve as a challenging and general purpose NLU benchmark. Consequently, \citet{N18-1101} introduced \textsc{MNLI} containing $433$K sentence pairs where the premises are extracted from a diverse number of sources such as face-to-face conversations, travel guides,
and the 9/11 event. Apart from the premise sources, both \textsc{SNLI} and \textsc{MNLI} are constructed in a similar fashion and are the most popular NLI datasets in the recent years.

\setlength\dashlinedash{0.2pt}
\setlength\dashlinegap{1.5pt}
\begin{table*}[t]

\centering
\small

\scalebox{0.95}{
  \begin{tabular} { p{5em} p{18em}  p{17.5em} p{5em}}
    \toprule
   \multicolumn{1}{l}{\rule{0pt}{1ex}\textbf{Domain}} & \multicolumn{1}{c}{\rule{0pt}{1ex}\textbf{First Sentence}} & \multicolumn{1}{c}{\textbf{Second Sentence}}&
   \multicolumn{1}{c}{\rule{0pt}{1ex}\textbf{Class}}\\
    \hline
    \textsc{NeurIPS}  & A number of psychological studies have suggested that our brains indeed perform causal inference as an ideal observer (e.g., [10, 12–14]). & \sout{However,} it has been challenging to come up with a simple and biologically plausible neural implementation for causal inference. & \textsc{Contrasting}\\
    \hdashline
    \textsc{Networks}  & Researchers found out that the inhomogeneity in the spatio-temporal distribution of the data traffic leads to extremely insufficient utilization of network resources. & \sout{Thus,} it is important to fundamentally understand this distribution to help us make better resource planning or introduce new management tools such as time-dependent pricing to reduce the congestion. & \textsc{Reasoning} \\
    \hdashline
    \textsc{Hardware}  & Scaling PCM in deep sub-micron regime faces non-negligible inter-cell thermal interference during programming, referred to as write disturbance (WD) phenomenon. & 
    \sout{That is,} the heat generated for writing one cell may disseminate beyond this cell and disturb the resistance states of its neighboring cells. & \textsc{Entailment}\\
    \hdashline
    \textsc{Security \& Privacy} & Following Google's best practices for developing secure apps, the password database is saved in the app data folder, which should be accessible only to the app itself. & this defines a hierarchical relationship between domains where the bounded domain cannot have more permissions than its bounding domain (the parent). & \textsc{Neutral} \\
    \hdashline

    \textsc{Software \& its Engineering} & If the delete operation is complex, then it advances to the discovery mode after which it will advance to the cleanup mode. & \sout{On the other hand,} if it is simple, then it directly advances to the cleanup mode (and skips the discovery mode). & \textsc{Contrasting} \\
    \bottomrule
    
  \end{tabular}}
  \vspace{-2mm}
  \caption{\small Examples of sentence pairs from \textsc{MSciNLI} extracted from different domains. The linking phrases at the beginning of the second sentence (strikethrough text in the table) are deleted after extracting the sentence pairs and assigning the labels.}

    \label{table:class_examples}
    \vspace{-2mm}
\end{table*}

Other NLI datasets include \textsc{QNLI} \cite{wang-etal-2018-glue}---derived from the SQuaD \cite{rajpurkar-etal-2016-squad} question-answering dataset; \textsc{XNLI} \cite{conneau2018xnli}---a cross lingual evaluation corpus derived by translating examples from \textsc{MNLI}; \textsc{ANLI} \cite{nie-etal-2020-adversarial}---constructed in an iterative adversarial fashion to reduce spurious patterns where human annotators develop examples that can cause the model to make errors in each iteration; \textsc{SciTail} \cite{khot2018scitail}---derived from a school level science question-answer corpus in which the sentence pairs are classified into two classes: entailment or not-entailment. These datasets have also seen wide applications both as NLU benchmarks and to improve other downstream NLP tasks. However, none of these datasets contains sentences from scientific text that is found in research articles. Moreover, the classes in these datasets are not sufficient to study the inter-sentence inferences and complexities that occur frequently in scientific text. 

Thus, to capture both the particularities in scientific text and provide coverage to the frequently occurring inter-sentence semantic relations, \citet{sadat-caragea-2022-scinli} introduced \textsc{SciNLI}. The sentence pairs in \textsc{SciNLI} were extracted from papers published in the ACL anthology \cite{radev-etal-2009-acl} using distant supervision based on different linking phrases. Given that \textsc{SciNLI} was derived from a single data source (ACL), it also lacks the necessary diversity in the data. Therefore, with a similar motivation behind constructing \textsc{MNLI}---to extend SNLI to multiple domains, we propose \textsc{MSciNLI}, the first diverse benchmark for scientific NLI, to extend \textsc{SciNLI} to multiple domains. The availability of multiple domains in \textsc{MSciNLI} enables the evaluation of the models' generalization ability under domain shift.

Recently, LLMs have demonstrated near human performance in many NLP tasks including NLI. For example, \citet{zhong2023can} reported that ChatGPT\footnote{\url{https://chat.openai.com/}} shows a zero-shot accuracy of $88\%$ on \textsc{RTE} and $89.3\%$ on the matched test set of \textsc{MNLI}. Thus, developing benchmark tasks and datasets which are challenging for even powerful LLMs is paramount. While the primary goal of our dataset is to introduce diversity in scientific NLI, because of the complex reasoning and inference required to predict the semantic relation between a pair of sentences from scientific text, it can serve as a challenging benchmark even for powerful LLMs. 

\section{MSciNLI: A Multi-Domain Scientific NLI Benchmark}

In this section, we describe the data sources for \textsc{MSciNLI}, its construction process and statistics.

\vspace{-2mm}
\subsection{Data Sources}
We derive \textsc{MSciNLI} from the papers published in four categories of the ACM digital library\footnote{\url{https://dl.acm.org/}} --- `Hardware', `Networks', `Software and its Engineering', `Security and Privacy' and the papers published in the NeurIPS\footnote{\url{https://papers.nips.cc/}} conference. Table \ref{table:class_examples} shows examples of sentence pairs extracted from our five domains. Further details on our data sources (e.g., publication years of the papers) are available in Appendix \ref{appendix:data_sources}.

\subsection{Data Extraction and Automatic Distant Supervision Labeling}

We closely follow the data extraction and automatic labeling procedure based on distant supervision proposed by \citet{sadat-caragea-2022-scinli}. Specifically, we use linking phrases between sentences (e.g., ``Therefore'', ``Thus'', ``In contrast'', etc.) to automatically annotate a large (potentially noisy) training set with the NLI relations. The complete list of linking phrases and their mapping to the NLI relations are presented in Appendix \ref{appendix:list_of_phrases}. The procedure is detailed below.

For the \textsc{Entailment}, \textsc{Contrasting}, and \textsc{Reasoning} classes, we extract adjacent sentence pairs from the papers collected from our five domains such that the second sentence starts with a linking phrase. For each extracted sentence pair, the relation corresponding to the linking phrase at the beginning of the second sentence is assigned as its class label. For example, if the second sentence starts with `Therefore' or `Consequently', the example is labeled as \textsc{Reasoning}. Note that the linking phrase is removed from the second sentence after assigning the label to prevent the models from predicting the label by simply learning a superficial correlation between the linking phrase and the label and without actually learning the semantic relation.

For the \textsc{Neutral} class, we construct the sentence pairs by extracting both sentences in the pair from the same paper using three approaches as follows: a) two random sentences that do not begin with any linking phrase are paired together; b) a random sentence which does not begin with any linking phrase is chosen as the first sentence and is paired with the second sentence of a random pair that belongs to one of the other three classes; c) a random sentence which does not begin with any linking phrase is chosen as the second sentence and is paired with the first sentence of a random pair that belongs to one of the other three classes. 

After extracting the sentence pairs for all four classes, we randomly split them at paper level into train, test and development sets (to ensure that the sentence pairs extracted from a certain paper end up in a single set). We directly use the automatically annotated examples for training the models. However, our use of distant supervision during the construction of the training set may introduce label noise when the relation between a pair of sentences is not accurately captured by the linking phrase. Therefore, to ensure a realistic evaluation, we employ human annotators to manually annotate the sentence pairs in the test and dev sets with one of the four scientific NLI relations as described below.

\vspace{-1mm}
\subsection{Multi-domain Scientific NLI Test and Development Set Creation}
\vspace{-1mm}
\label{sec:manual_annotation}

\begin{table*}[t]
\centering
\small

\scalebox{1.00}{
  \begin{tabular}{ l r r r r r r r r r }
    \toprule
      &  \multicolumn{3}{c}{\bf \#Examples} & \multicolumn{2}{c}{\bf \#Words} & \multicolumn{2}{c}{\bf `S' parser} & \multicolumn{1}{c}{\bf Word} \\
       \cmidrule(lr){2-4}  \cmidrule(lr){5-6}  \cmidrule(lr){7-8}
   {\bf Domain } & {\bf Train}     & {\bf Dev}   & {\bf Test} & {\bf Prem.} & {\bf Hyp.} & {\bf Prem.} & {\bf Hyp.} & {\bf Overlap} & {\bf Agrmt.}  \\ 
   \midrule
   \textsc{SciNLI (ACL)} & 101,412 & 2,000 & 4,000 & 27.38 & 25.93 & 96.8\% & 96.7\% & 30.06\% & 85.8\%\\  
    \midrule 
   \textsc{Hardware} & 25,464 & 200 & 800 & 26.10 & 24.59 & 94.3\% & 94.5\% & 30.52\% & 84.6\%\\
    \textsc{Networks} & 25,464 & 200 &  800 & 26.37 & 25.01 & 93.9\% & 93.7\% & 30.17\% & 90.5\%\\
    \textsc{Software \& its Engineering} & 25,464 & 200 & 800 & 25.80 & 24.51 & 93.9\% & 94.1\% & 29.83\% & 86.5\%\\
    \textsc{Security \& Privacy} & 25,464 & 200 & 800 & 26.14 & 24.50 & 94.0\% & 94.2\% & 29.91\% & 90.4\%\\
   \textsc{NeurIPS} & 25,464 & 200 & 800 & 29.80 & 29.66 & 96.0\% & 95.1\% & 31.04\% & 88.5\%\\
    
\midrule
\textsc{MSciNLI} Overall & 127,320 & 1,000 & 4,000 & 26.84 & 25.85 & 94.4\% & 94.3\% & 30.29\% & 88.0\%\\
    \bottomrule
  \end{tabular}}
  \caption{\small Comparison of key statistics of \textsc{MSciNLI} with \textsc{SciNLI}.}
  \vspace{-4.5mm}
    \label{table:data_stat}
\end{table*}

Three expert annotators (see Appendix \ref{annotator_details} for more details about the annotators and the instructions) are employed to annotate the test and dev sets of \textsc{MSciNLI}. Specifically, a random subset (balanced over the classes) of sentence pairs 
from the test and dev sets are given to the three annotators who are instructed to annotate their labels (the relation between the sentences) based only on the context available in the two sentences in each example. If the annotators are unable to determine the label based on the two sentences of a pair, they mark it as unclear. We assign a gold label to each example based on the majority vote from the annotators. In rare cases ($\approx 3\%$) where there is no consensus among the annotators for an example, we do not assign a gold label. The examples for which there is a match between the gold label and the automatically assigned label (based on linking phrases) are included in their respective split and the rest are filtered out.

For each domain, we continue sampling random subsets (without replacement) of examples and manually annotate them until we have at least $800$ clean examples ($200$ from each class) in the test set and $200$ clean examples ($50$ from each class) in the dev set. In total, we annotate $6,992$ examples (all domains combined), among which $6,153$ have an agreement between the gold label and the automatically assigned label. That is, the overall agreement rate for \textsc{MSciNLI} is $88.0\%$. Moreover, we find a Fleiss-k score of $70.51\%$ for \textsc{MSciNLI} indicating substantial agreement among the annotators \cite{landis1977measurement}. 

\vspace{-1mm}
\paragraph{Data Balancing}
To ensure equal representation, the number of examples per class in each domain are downsampled to a size of $200$ and $50$ in the test and dev set, respectively. Consequently, we end up with a combined (over the domains) test and dev sets of $4000$ and $1000$ examples, respectively (balanced over the classes and domains). We balance the training set by using a similar procedure. 

\vspace{-1mm}
\subsection{Data Statistics}
\vspace{-1mm}
We show a comparison of key statistics of our dataset with the \textsc{SciNLI} dataset in Table \ref{table:data_stat}.

\vspace{-2mm}
\paragraph{Dataset Size}
We can see that the total number of examples (<premise, hypothesis> pairs) in \textsc{MSciNLI} is higher than that in \textsc{SciNLI}, the only NLI dataset over scientific text. Moreover, each domain in \textsc{MSciNLI} has a large number of examples in the training set which enables exploration of NLI in-domain as well as across domains.

\vspace{-1mm}
\paragraph{Sentence Parses} Similar to \textsc{SciNLI}, we use the Stanford PCFG Parser (3.5.2) \cite{klein-manning-2003-accurate} to parse the sentences in our dataset. We can see in Table \ref{table:data_stat} that $\approx 94\%$ of the sentences in \textsc{MSciNLI} have an `S' root showing that most sentences in our dataset are syntactically complete.

\vspace{-1mm}
\paragraph{Token Overlap}
 The percentage of word overlap between the premise and hypothesis in each pair in \textsc{MSciNLI} is also low and close to that of \textsc{SciNLI} as shown in Table \ref{table:data_stat}. Thus, like \textsc{SciNLI}, our \textsc{MSciNLI} dataset is also less vulnerable to surface level lexical cues.

\section{\textsc{MSciNLI} Evaluation}
Our main experiments for evaluating \textsc{MSciNLI} consists of three stages. First, we evaluate its difficulty by experimenting with a BiLSTM model (\S \ref{sec:difficulty_eval}). Next, we establish strong baselines on \textsc{MSciNLI} with four Pre-trained Language Models (PLMs) and two Large Language Models (LLMs), and compare them with human performance (\S \ref{sec:baselines}). Finally, we analyze our best performing baseline by investigating its performance when it is fine-tuned on various subsets of the training set and its performance under domain shift 
(\S \ref{sec:analysis}). Our implementation details are given in Appendix \ref{appendix:implementation_details}. Additional experiments on the impact of dataset size and diversity in model training; performance of another LLM; spurious correlations \cite{gururangan-etal-2018-annotation}; and class-wise performances of the baselines are shown in Appendix \ref{appendix:additional_results}.

\subsection{Difficulty Evaluation}
\label{sec:difficulty_eval}
\paragraph{BiLSTM Model} The architecture of this model (described in Appendix \ref{appendix:implementation_details}) is similar to the BiLSTM model adopted by \citet{williams-etal-2018-broad}. We can see a comparison of the performance of this model on \textsc{MSciNLI} and \textsc{SciNLI} in Table \ref{table:lstm_results_clean}. We observe the following:

\vspace{-1mm}
\paragraph{\textsc{MSciNLI} is more challenging than \textsc{SciNLI}.} We can see that the Macro F1 of the BiLSTM model for \textsc{SciNLI} is $61.12\%$ whereas it is only $54.40\%$ for \textsc{MSciNLI} (the model is trained on the combined \textsc{MSciNLI} training set). These results indicate that \textsc{MSciNLI} presents a broader range of challenges for the model compared with \textsc{SciNLI},
making the scientific NLI task more difficult.

\subsection{Baselines}
Here, we describe the baseline models for \textsc{MSciNLI} and discuss their performance.
\label{sec:baselines}
\subsubsection{PLM Baselines}
\label{sec:PLM_baselines}

We fine-tune the base variants of the following PLMs on the combined \textsc{MSciNLI} training set: \textbf{\textsc{BERT}} \cite{devlin-etal-2019-bert}; \textbf{\textsc{SciBERT}} \cite{beltagy-etal-2019-scibert}; (c) \textbf{\textsc{RoBERTa}} \cite{liu2019roberta}; and (d) \textbf{\textsc{XLNet}} \cite{yang2019xlnet}. We run each experiment with the PLM baselines three times with different random seeds and report the average and standard deviation of their domain-wise and overall Macro F1 scores in Table \ref{table:baseline_results}. Our findings are described below.

\paragraph{Domain specific pre-training helps improve the performance.} We can see that \textsc{SciBERT} shows a better performance than \textsc{BERT} in all domains. Note that 
\textsc{SciBERT} does not address any weaknesses of \textsc{BERT} and is trained using the same procedure as \textsc{BERT}, except \textsc{SciBERT} exclusively uses scientific text for pre-training whereas \textsc{BERT} is trained on the BookCorpus and Wikipedia. Thus, pre-training on scientific documents helps improve the performance of scientific NLI.

\setlength\dashlinedash{0.2pt}
\setlength\dashlinegap{1.5pt}
\setlength\arrayrulewidth{0.3pt}
\begin{table}[t]
\centering
\small

\scalebox{1.00}{

  \begin{tabular}{l c c }
    \toprule
{\bf Dataset} & {\bf F1}     & {\bf Acc} \\ 
   \midrule
    \textsc{SciNLI (ACL)} & $61.12$ & $61.32$ \\
    \hdashline
    \textsc{MSciNLI} & & \\
    \mbox{{\color{white}{xxx}}-}Hardware & $53.61$ & $53.87$ \\
    \mbox{{\color{white}{xxx}}-}Networks & $54.78$ & $54.95$ \\
    \mbox{{\color{white}{xxx}}-}Software \& its Engineering & $51.96$ & $52.20$ \\
    \mbox{{\color{white}{xxx}}-}Security \& Privacy & $52.18$ & $52.62$ \\
    \mbox{{\color{white}{xxx}}-}NeurIPS & $59.19$ & $59.41$ \\
    \hdashline
     \mbox{{\color{white}{xxx}}-}\textbf{Overall} & $54.40$ & $54.61$ \\
    \bottomrule
  \end{tabular}}
\vspace{-3mm}
  \caption{\small The Macro F1 (\%) and Accuracy (\%) of the BiLSTM model on \textsc{SciNLI} and \textsc{MSciNLI}.
  }
\vspace{-2mm}
    \label{table:lstm_results_clean}
    
\end{table}

\vspace{-1mm}
\paragraph{``Robust'' pre-training leads to better performance.} Both \textsc{RoBERTa} and \textsc{XLNet} are designed to address different weaknesses of \textsc{BERT}. \textsc{RoBERTa} focuses on optimizing the model in a more robust manner during pre-training while \textsc{XLNet} aims at incorporating auto-regressive nature of natural language without removing bi-directional context. Both of these models substantially outperform \textsc{BERT} in all domains and \textsc{RoBERTa} consistently outperforms \textsc{XLNet}. We can also observe that \textsc{RoBERTa} leads to even better performance compared with SciBERT in most cases.

\begin{table*}[t]
\centering
\small

\scalebox{1.00}{
  \begin{tabular}{l c c c c c c}
    \toprule

{\textsc{Model}} &  {\textsc{Hardware}} &  {\textsc{Networks}} & {\textsc{SWE}} & {\textsc{Security}} & {\textsc{NeurIPS}} & {\textsc{Overall}}\\
    
    \midrule

    \textsc{BERT} & $72.89 \pm 0.1$ & $74.10 \pm 1.3$ & $71.37 \pm 0.3$ & $72.38 \pm 2.5$ & $75.46 \pm 0.8$ & $73.24 \pm 0.8$\\
    \textsc{SciBERT} & $\underline {75.91 \pm 0.1}$ & ${\bf 76.51 \pm 0.5}$ & $\underline {75.28 \pm 1.1}$ & $\underline {75.94 \pm 0.4}$ & ${\bf 78.78 \pm 0.1}$ & ${\underline {76.48 \pm 0.4}}$\\
    \textsc{XLNet} & $75.59 \pm 0.5$ & $75.25 \pm 0.1$ & $73.98 \pm 0.6$ & $75.09 \pm 0.8$ & $77.64 \pm 1.0$ & $75.51 \pm 0.3$\\
    \textsc{RoBERTa} & ${\bf 77.79^{\$} \pm 0.2}$ & $\underline {75.45 \pm 1.5}$ & ${\bf 77.10^{\#} \pm 0.7}$ & ${\bf 77.71^{\$} \pm 0.2}$ & $\underline {78.04 \pm 0.8}$ & ${\bf 77.21^{\#} \pm 0.3}$\\
    \midrule

  \end{tabular}}
\vspace{-2mm}
   \caption{\small Macro F1 scores (\%) of the PLM baselines on different domains. Here, \textsc{SWE}: Software \& its Engineering and \textsc{Security}: Security \& Privacy. $^\#$ and $^\$$ indicate statistically significant improvement by \textsc{RoBERTa} over \textsc{XLNet} and over both \textsc{SciBERT} and \textsc{XLNet}, respectively according to a paired t-test with $p < 0.05$. Best performance is shown in {\bf bold}, and second best is {\underline{underlined}}. 
  }
  
    \label{table:baseline_results}
\end{table*}

\begin{table*}[t]
\centering
\small

\scalebox{1.00}{
  \begin{tabular}{l l c c c c c c}
    \toprule

{\textsc{Model}} &  {\textsc{Prompt}} & {\textsc{Hardware}} &  {\textsc{Networks}} & {\textsc{SWE}} & {\textsc{Security}} & {\textsc{NeurIPS}} & {\textsc{Overall}}\\

    \midrule
    \textbf{\textsc{Llama-2}} & \textsc{Prompt - 1}$_{zs}$ & $20.31$ & $21.34$ & $19.77$ & $21.36$ & $18.92$ & $20.41$\\
    & \textsc{Prompt - 2}$_{zs}$ & $18.23$ & $20.60$ & $21.26$ & $19.87$ & $17.62$ & $19.53$\\
    & \textsc{Prompt - 3}$_{zs}$ & $30.27$ & $32.64$ & $30.49$ & $30.16$ & $27.58$ & $30.36$\\
    \hdashline
     & \textsc{Prompt - 1$_{fs}$} & $24.42$ & $26.69$ & $27.75$ & $28.84$ & $22.98$ & $26.21$\\
    & \textsc{Prompt - 2$_{fs}$} & $37.49$ & $38.27$ & $34.25$ & $36.32$ & $35.26$ & $36.39$\\
    & \textsc{Prompt - 3$_{fs}$} & ${\bf 53.41}$ & ${\bf 51.38}$ & ${\bf 50.54}$ & ${\bf 52.75}$ & ${\bf 50.38}$ & ${\bf 51.77}$\\
    \midrule

    \textbf{\textsc{Mistral}} & \textsc{Prompt - 1}$_{zs}$ & $21.72$ & $21.48$ & $19.87$ & $22.77$ & $21.36$ & $21.43$\\
    & \textsc{Prompt - 2}$_{zs}$ & $34.54$ & $32.95$ & $32.5$ & $33.51$ & $34.71$ & $33.66$\\
    & \textsc{Prompt - 3}$_{zs}$ & $34.64$ & $33.68$ & $34.14$ & $36.00$ & $34.78$ & $35.00$\\
    \hdashline
     & \textsc{Prompt - 1$_{fs}$} & $\underline {48.21}$ & $\underline {42.50}$ & $\underline {45.68}$ & $\underline {44.40}$ & $\underline {45.98}$ & $\underline {45.49}$\\
    & \textsc{Prompt - 2$_{fs}$} & $39.83$ & $38.71$ & $35.45$ & $36.70$ & $36.30$ & $37.55$\\
    & \textsc{Prompt - 3$_{fs}$} & $30.75$ & $31.17$ & $31.23$ & $34.38$ & $21.92$ & $30.23$\\
    \midrule
    
  \end{tabular}}
\vspace{-3mm}
   \caption{\small Macro F1 scores (\%) of the LLM baselines on different domains. Here, \textsc{SWE}: Software \& its Engineering and \textsc{Security}: Security \& Privacy. Best performance is shown in {\bf bold}, and second best is {\underline{underlined}}.  
  }
  
    \label{table:baseline_results_LLMs}
\vspace{-3mm}
\end{table*}

\subsubsection{LLM Baselines}
\label{sec:LLM_baselines}
We experiment with two LLMs as baselines for our dataset: (a) \textbf{\textsc{Llama-2}} \cite{touvron2023llama} and (b) \textbf{\textsc{Mistral}} \cite{jiang2023mistral}. More specifically, we use the \textit{Llama-2-13b-chat-hf} and \textit{
Mistral-7B-Instruct-v0.1} variants of \textsc{Llama-2} and \textsc{Mistral}, containing $13$ billion and $7$ billion parameters, respectively. Both of these models are chosen because of their success in many NLP tasks that require complex reasoning and problem solving (e.g., the MMLU benchmark \cite{hendrycks2021measuring}). 

We construct $3$ multiple-choice question templates for the scientific NLI task to be used for prompting the LLMs:
\begin{itemize}
\vspace{-1mm}
    \item {\bf \textsc{Prompt - 1}}: this prompt asks the LLMs to predict the class given a sentence pair with the four class names as the choices.
\vspace{-2mm}
    \item {\bf \textsc{Prompt - 2}}:  to provide further context to the LLMs about the scientific NLI task, this prompt first defines the scientific NLI classes and then poses the question to predict the class with the class names as the choices. 
\vspace{-2mm}
\item {\bf \textsc{Prompt - 3}}: instead of providing the definitions of the classes first and then asking a question with the class names as the choices, this prompt directly uses the class definitions as the choices.

\end{itemize}
\vspace{-2mm}
The three prompt templates can be seen in 
Appendix \ref{appendix:prompts}. We evaluate the performance of the LLMs in two settings: a) zero-shot: no input-output exemplars are shown to the model; b) few-shot: four input-human-annotated output exemplars (one for each class) are pre-pended to the prompt to evaluate the LLMs' in-context learning \cite{brown2020language} ability for scientific NLI. The zero-shot and few-shot versions of each prompt $i$ is denoted as \textsc{Prompt - $i$}$_{zs}$, and \textsc{Prompt - $i$}$_{fs}$, respectively.

We employ a greedy decoding strategy for all of our LLM based experiments and report the domain-wise and the overall Macro F1 scores of each experiment in Table \ref{table:baseline_results_LLMs}. We find the following:

\vspace{-1mm}
\paragraph{\textsc{Llama-2} performs better than \textsc{Mistral}.} We can see that \textsc{Llama-2} with \textsc{Prompt - 3$_{fs}$} shows the best performance among all of our LLMs with a Macro F1 of $51.77\%$. This is $6.28\%$ higher than the best performance shown by \textsc{Mistral} with \textsc{{Prompt - 1$_{fs}$}}. Thus, \textsc{Llama-2} with its $13$B parameters has more complex reasoning capability compared to \textsc{Mistral} with its $7$B parameters.

\paragraph{Using class-definitions as choices in the prompt and the few-shot prompt variants improve the performance.} We can see that the performance of both LLMs are generally better when we use \textsc{Prompt - 3}. This indicates that using the class definitions as the potential choices in the multiple-choice question is more suitable for the models, resulting in better performance. We can also see that the few-shot variants of the prompts generally outperform their zero-shot counterparts. Thus, both LLMs are capable of in-context learning and providing few examples can boost their performance.

\paragraph{Scientific NLI is highly challenging for state-of-the-art LLMs} Despite the promising few-shot performance, based on the results in Table \ref{table:baseline_results_LLMs}, it is evident that the task of scientific NLI is highly challenging even for powerful LLMs. Therefore, our dataset along with \textsc{SciNLI} can serve as a challenging evaluation benchmark for LLMs.

\setlength\dashlinedash{0.2pt}
\setlength\dashlinegap{1.5pt}
\setlength\arrayrulewidth{0.3pt}
\begin{table}[t]
\centering
\small

\scalebox{1.00}{

  \begin{tabular}{l c c}
    \toprule
Method & {\bf Macro F1}   & {\bf Accuracy}\\ 
   \midrule
    \textsc{RoBERTa} &  $77.21 \pm 0.30$ &  $77.42 \pm 0.30$\\
    \textsc{Llama-2} &  $51.77 \pm 0.00$ &  $51.10 \pm 
 0.00$\\
    \hdashline
    \textsc{Human - E (est.)} &  $89.33 \pm 1.18$ &  $89.10 \pm 1.10$\\
    \textsc{Human - NE (est.)} &  $79.78 \pm 4.43$ &  $79.49 \pm 4.84$\\
    
    \bottomrule
  \end{tabular}}

  \caption{\small Comparison of \textit{estimated} human expert and non-expert performances with \textsc{RoBERTa} and \textsc{Llama-2} (with \textsc{Prompt - 3}) on the \textsc{MSciNLI} test set. Here, E: expert, NE: non-expert.  
  }
\vspace{-4mm}
    \label{table:human_performance}
    
\end{table}

\subsubsection{Human Performance}

\label{sec:human_performance}

We hire three expert annotators (with relevant domain-specific background) and three non-expert annotators (with no background in any of the five domains) to evaluate the human performance on \textsc{MSciNLI}. Note that these expert and non-expert annotators are not involved in our dataset construction process (see Appendix \ref{annotator_details} for more details). Following other popular benchmarks (e.g., \textsc{SuperGLUE} \cite{NEURIPS2019_4496bf24}), we \textit{estimate} the human performance by re-annotating a small randomly sampled subset of our test set. 
Each example in the subset is re-annotated by $3$ expert and $3$ non-expert annotators following the same data annotation procedure described in Section \ref{sec:manual_annotation}. We report the average and the standard deviation of the expert and non-expert performances (Macro F1) on this subset, and compare them with the best performing PLM baseline, \textsc{RoBERTa}, and the best performing LLM baseline, \textsc{Llama-2} with \textsc{Prompt - 3$_{fs}$} in Table \ref{table:human_performance}. Our findings are described below:

\paragraph{Experts outperform non-experts, and a substantial gap exists between model performance and human expert performance.} As expected, expert annotators with the relevant domain-specific knowledge substantially outperform the non-expert annotators. Despite the lower performance by the non-experts (compared with experts), we can see that they still outperform our baselines. Furthermore, the performance by the experts is significantly higher than both \textsc{RoBERTa} and \textsc{Llama-2}. Therefore, there is a substantial headroom for improving the models' performance which can foster future research on scientific NLI.

\vspace{-1mm}
\subsection{Analysis}

In this section, first, we diagnose the \textsc{MSciNLI} training set by fine-tuning separate models using different training subsets selected by performing data cartography \cite{swayamdipta-etal-2020-dataset} (\S\ref{sec:cartography}). Next, we study the model behavior under domain shift at test time (\S\ref{sec:ood_performance}). Finally, we perform cross-dataset experiments where we analyze the performance of models fine-tuned on \textsc{SciNLI}, \textsc{MSciNLI}, and their combination (\S\ref{sec:cross_dataset}). We choose our best performing baseline model, \textsc{RoBERTa} for these experiments.

\label{sec:analysis}

\vspace{-1mm}
\subsubsection{Data Cartography Experiments}
\label{sec:cartography}
\vspace{-1mm}
We perform a data cartography of \textsc{MSciNLI} to characterize each example in the training set using two metrics --- \textit{confidence} and \textit{variability}. Based on this characterization, inspired by \cite{swayamdipta-etal-2020-dataset}, first, we fine-tune three different \textsc{RoBERTa} models using the following subsets of the training set: 1) $33\%$ \textit{easy-to-learn} --- examples with high confidence; 2) $33\%$ \textit{hard-to-learn} --- examples with low confidence; 3) $33\%$ \textit{ambiguous} --- examples with high variability (the detailed method used for selecting these subsets of training examples is available in Appendix \ref{appendix: training_dynamics}). In addition, to further understand the effect of \textit{hard-to-learn} examples in model training, we fine-tune two other models using the full training set \textbf{minus} --- 1) top $25\%$ \textit{hard-to-learn} ($25\%$ examples with lowest confidence) and 2) top $5\%$ \textit{hard-to-learn} examples ($5\%$ examples with lowest confidence), denoted as `$100\% - $ top $25\%$ \textit{hard}', and `$100\% - $ top $5\%$ \textit{hard}', respectively. The results are shown in Table \ref{table:region_wise_results}. We find the following.

\vspace{-1mm}

\setlength\dashlinedash{0.2pt}
\setlength\dashlinegap{1.5pt}
\setlength\arrayrulewidth{0.3pt}
\begin{table}[t]
\centering
\small

\scalebox{1.00}{

  \begin{tabular}{l c c}
    \toprule
{\bf Data Subset} & {\bf Macro F1}   & {\bf Accuracy}\\ 
   \midrule
    $100\%$ &  $77.21 \pm 0.30$ &  $77.42 \pm 0.30$\\
    \hdashline
    $33\%$ \textit{easy-to-learn} & $73.71^{*} \pm 1.40$ &  $73.74^{*} \pm 1.43$\\
    $33\%$ \textit{hard-to-learn} & $34.11^{*} \pm 5.65$ &  $37.99^{*} \pm 1.53$\\
    $33\%$ \textit{ambiguous} & $75.65^{*} \pm 0.27$ &  $75.57^{*} \pm 0.26$\\
    \hdashline
    $100\% -$ top $25\%$ \textit{hard} & $76.60 \pm 0.65$ &  $76.64 \pm 0.66$\\
    $100\% -$ top $5\%$ \textit{hard} & $77.47 \pm 0.23$ &  $77.44 \pm 0.28$\\
    
    \bottomrule
  \end{tabular}}
  \vspace{-2mm}
  \caption{\small The Macro F1 (\%) and Accuracy (\%) of \textsc{RoBERTa} fine-tuned on different subsets of \textsc{MSciNLI} training set. $^*$ indicates a statistically significant difference with the performance of the model trained on $100\%$ data according to a paired t-test with $p < 0.05$.
  }
\vspace{-4mm}
    \label{table:region_wise_results}
    
\end{table}

\paragraph{Ambiguous examples yield stronger models while the full training set yields better performance.}

We can see that the model fine-tuned on the $33\%$ \textit{ambiguous} examples shows the best performance among the $33\%$ subsets. Therefore, `ambiguousness' in the training examples helps train stronger scientific NLI models. Despite the strong performance shown by $33\%$ \textit{ambiguous}, its Macro F1 is still lower than the $100\%$ of the training set. Furthermore, although $33\%$ \textit{hard-to-learn} shows a poor performance ($34.11\%$ in Macro F1), removing a percentage of them (e.g., $25\%$, and $5\%$ in the bottom block of Table \ref{table:region_wise_results}) from the $100\%$ of the training set does not result in any statistically significant difference in performance compared with $100\%$. Therefore, all examples in the training set are useful for training the most optimal model.

\vspace{-1mm}
\subsubsection{Out-of-domain Experiments}
\vspace{-1mm}
\label{sec:ood_performance}
Here, we train \textsc{RoBERTa} on one domain and test it on another domain (out-of-domain) and contrast it with the \textsc{RoBERTa} trained and tested on the same domain (in-domain). In addition to the five domains in \textsc{MSciNLI}, we also experiment with the \textsc{ACL} domain from \textsc{SciNLI}. For a fair comparison with the other domains, we downsample the training set from \textsc{SciNLI} to the same size as that of the other domains and denote it as \textsc{ACL - small}. Both in-domain (ID) and out-of-domain (OOD) results are shown in Table \ref{table:OOD_results}. Our findings are described below:

\setlength\dashlinedash{0.2pt}
\setlength\dashlinegap{1.5pt}
\setlength\arrayrulewidth{0.3pt}
\begin{table*}[t]
\centering
\small

\vspace{2mm}
\scalebox{1.00}{
  \begin{tabular}{l c c c c c c}
    \toprule
 {\bf \backslashbox{Train}{Test}} &  {\bf \textsc{Hardware}} & {\bf \textsc{Networks}} & {\bf \textsc{SWE}} & {\bf \textsc{Security}} & {\bf \textsc{NeurIPS}} & {\bf \textsc{ACL}}\\
  \midrule
    {\bf \textsc{Hardware}} & \cellcolor{light-gray}$74.93 \pm 1.4$ & $73.11 \pm 1.2$ & $74.24 \pm 0.2$ & $72.98 \pm 2.3$ & $73.97 \pm 0.7$ & $72.40 \pm 0.8$\\
     {\bf \textsc{Networks}} & $75.04 \pm 1.3$ & \cellcolor{light-gray}$73.31 \pm 1.7$ & $73.29 \pm 0.5$ & $73.44 \pm 1.0$ & $74.61 \pm 1.1$ & $72.72 \pm 1.0$\\
     {\bf \textsc{SWE}} & $73.60 \pm 1.1$ & $71.25 \pm 0.8$ & \cellcolor{light-gray}$74.44 \pm 0.5$ & $73.24 \pm 1.4$ & $75.31 \pm 1.4$ & $73.06 \pm 1.8$\\
     {\bf \textsc{Security}} & $72.69 \pm 2.5$ & $70.94 \pm 1.8$ & $74.14 \pm 1.7$ & \cellcolor{light-gray}$74.45 \pm 2.5$ & $73.12 \pm 1.6$ & $72.85 \pm 1.3$\\
     {\bf \textsc{NeurIPS}} & $73.64 \pm 0.8$ & $71.94 \pm 1.1$ & $71.76 \pm 1.2$ & $71.62 \pm 0.8$ & \cellcolor{light-gray}$76.02 \pm 1.1$ & $74.15 \pm 0.8$\\
     {\bf \textsc{ACL - small}} & $74.29 \pm 0.1$ & $71.81 \pm 0.8$ & $73.44 \pm 1.4$ & $73.38 \pm 1.1$ & $74.95 \pm 1.9$ & \cellcolor{light-gray}$75.30 \pm 0.5$\\

    \bottomrule
  \end{tabular}}
  \vspace{-3mm}
\caption{\small ID and OOD Macro F1 (\%) of \textsc{RoBERTa} models trained on different domains. ID performance is shown in \colorbox{light-gray}{gray.} Here, \textsc{SWE: Software \& its Engineering} and \textsc{Security: Security \& Privacy}.
  }
  \vspace{-2mm}
  
    \label{table:OOD_results}

\end{table*}

\setlength\dashlinedash{0.2pt}
\setlength\dashlinegap{1.5pt}
\setlength\arrayrulewidth{0.3pt}
\begin{table}[t]
\centering
\small

\scalebox{1.00}{
  \begin{tabular}{l l c c}
    \toprule
    {\bf \backslashbox{Tr}{Te}} & {\bf SciNLI} & {\bf MSciNLI} & {\bf MSciNLI+}\\
   \toprule
    \textsc{SciNLI} & $78.08$ & $75.19$ & $76.63$\\
    \hdashline
    \textsc{MSciNLI} & $76.74$ & $77.21$ & $76.95$\\
    \hdashline
     \textsc{MSciNLI+ (s)} & $77.78$ & $77.37$ & $77.54$\\
    \textsc{MSciNLI+} & $\textbf{79.48}$ & $\textbf{78.07}$ & $\textbf{78.76}$\\
     
    \bottomrule
  \end{tabular}}
  \vspace{-3mm}
  \caption{\small Macro F1 scores of cross dataset experiments with \textsc{RoBERTa}. Here, Tr: Train, Te: Test.}
  \vspace{-5mm}
    
    \label{table:cross_dataset}

\end{table}

\vspace{-1mm}
\paragraph{The domain shift reduces the performance.} In general, for each domain, the ID model shows a higher performance than their OOD counterparts (see each column in Table \ref{table:OOD_results}). For example, the model fine-tuned on the \textsc{NeurIPS} training set shows a Macro F1 of $76.02\%$ when it is tested on \textsc{NeurIPS} as well. The performance sees a decline when the models trained on other domains are tested on \textsc{NeurIPS} (e.g., $74.61\%$ with the \textsc{Networks} model). This indicates that the sentence pairs in each domain exhibit unique linguistic characteristics which are better captured by a model trained on in-domain data.

\subsubsection{Cross-dataset Experiments}
\label{sec:cross_dataset}

For the cross-dataset experiments, we train four separate \textsc{RoBERTa} models on: 1) \textsc{SciNLI}, 2)  \textsc{MSciNLI}, 3) \textsc{MSciNLI+ (s)} - a combination of \textsc{MSciNLI} and \textsc{ACL - small}, and 4) \textsc{MSciNLI+} - a combination of \textsc{MSciNLI} and \textsc{SciNLI}. All four models are then evaluated using the separate \textsc{SciNLI} and \textsc{MSciNLI} test sets, and their combination i.e., the \textsc{MSciNLI+} test set. The results are reported in Table \ref{table:cross_dataset}. We also evaluate the models on the domain-wise test sets, and general domain NLI datasets, and report the results in Appendix \ref{sec:cross_dataset_additional}.

\looseness=-1
\paragraph{Diverse training data leads to robust models.} The performance sees a decline for both \textsc{SciNLI} and \textsc{MSciNLI} under `dataset-shift.' However, the model fine-tuned with \textsc{SciNLI} shows a higher drop in performance compared with the model fine-tuned with \textsc{MSciNLI} in the out-of-dataset setting. Specifically, the out-of-dataset Macro F1 of the model fine-tuned with \textsc{SciNLI} (when it is tested on \textsc{MSciNLI}) drops by $2.02\%$ from the in-dataset performance of \textsc{MSciNLI} ($77.21\%$). In contrast, the out-of-dataset Macro F1 of the model fine-tuned with \textsc{MSciNLI} (when it is tested on \textsc{SciNLI}) drops by only $1.34\%$ from the in-dataset performance of \textsc{SciNLI} ($78.08\%$). This indicates that the diversity in the data can train more robust scientific NLI models with stronger generalization capabilities.  

\paragraph{Combining the datasets yields the best performance.}
The best performance for both datasets and their combination is seen when the model is fine-tuned on \textsc{MSciNLI+}. Therefore, fine-tuning the model on a larger training set containing diverse examples yields better performance. We can see that the models trained on \textsc{MSciNLI+ (s)} show a lower performance than those trained on \textsc{MSciNLI+}. This is because \textsc{MSciNLI+ (s)} is smaller in size than \textsc{MSciNLI+}. However, due to the additional diversity introduced by  the ACL domain, \textsc{MSciNLI+ (s)} consistently outperforms \textsc{MSciNLI}. Thus, the benefit of combining the datasets holds for \textsc{MSciNLI+ (s)} as well.    

\setlength\dashlinedash{0.2pt}
\setlength\dashlinegap{1.5pt}
\setlength\arrayrulewidth{0.3pt}
\begin{table*}[t]
\centering
\small

\scalebox{1.00}{
  \begin{tabular}{l c c c c c}
    \toprule
&  \multicolumn{5}{c}{\bf Intermediate training data}\\
\cmidrule(lr){2-6}
    {\bf Dataset} & {\bf None} & {\bf MSciNLI+ (MLM)} & {\bf MNLI} & {\bf SciNLI} & {\bf MSciNLI+}\\
    
   \midrule
   \textsc{SciHTC} & $52.59$ & $48.95$ & $51.83$ & $51.83$  & $\textbf{53.47}$\\
     \textsc{Paper Field} & $73.66$ & $73.46$ & $73.64$ & $73.61$ & $\textbf{74.09}$\\
     \textsc{Acl-Arc}  & $69.57$ & $63.95$ & $59.73$ & $68.52$ & $\textbf{73.04}$\\

    \bottomrule
  \end{tabular}}
  \vspace{-1mm}
  \caption{\small Macro F1 ($\%$) of \textsc{RoBERTa} with intermediate task transfer using different NLI datasets.}  
\vspace{-2mm}

    \label{table:intermideate_training}

\end{table*}

\section{Scientific NLI as an Intermediate Task}
\vspace{-2mm}

Research \cite{martin2022facter, sadeghi2022fake} has shown that traditional NLI datasets (e.g., SNLI, MNLI) can aid in improving the performance of downstream NLP tasks. While the \textsc{SciNLI} dataset has already been used to improve sentence representation \cite{deka2022evidence}, it was used in conjunction with the traditional NLI datasets. In this section, we investigate whether the scientific NLI datasets by themselves can aid in improving the performance of downstream tasks in an intermediate task transfer setting \cite{pruksachatkun-etal-2020-intermediate}. 

To this end, first, a \textsc{RoBERTa} model (out-of-the-box pre-trained with a dynamic MLM objective) is fine-tuned on the downstream tasks. Next, we perform intermediate training of four separate out-of-the-box \textsc{RoBERTa} models with the following approaches before fine-tuning them on the downstream tasks: \textbf{1)} with a self-supervised dynamic MLM objective (with no information of the NLI classes) on \textsc{MSciNLI+}; \textbf{2)} with a supervised NLI objective using \textsc{MNLI}; \textbf{3)} with a supervised scientific NLI objective using \textsc{SciNLI}; \textbf{4)} with a supervised scientific NLI objective using \textsc{MSciNLI+}.

We experiment with the following downstream tasks: \textbf{\textsc{SciHTC}} \cite{sadat-caragea-2022-scihtc}, \textbf{\textsc{Paper Field}} \cite{beltagy-etal-2019-scibert}, and \textbf{\textsc{ACL-Arc}} \cite{jurgens2018measuring}. \textsc{SciHTC} and \textsc{Paper Field} are topic classification datasets for scientific papers and \textsc{ACL-Arc} is a citation intent classification dataset. Details on these  tasks and their labels are in Appendix \ref{appendix:intermediate_transfer}. The results for each downstream task are presented in Table \ref{table:intermideate_training}. We find that:

\paragraph{Scientific NLI can aid in improving the performance of downstream tasks.} As we can see from the table, intermediate training with an unsupervised MLM objective on \textsc{MSciNLI+} (MSciNLI+ (MLM) in the table) fails to improve the performance of the downstream tasks over the models which are fine-tuned without any intermediate training. In contrast, supervised intermediate training on \textsc{MSciNLI+} improves the performance of all datasets over all other models. This indicates that training a model further on the scientific NLI task can learn better and more relevant representations for the downstream tasks in the scientific domain. We can also see that supervised intermediate training on \textsc{MNLI} fails to show improvement for any of the downstream tasks. This illustrates the need for NLI datasets capturing the unique linguistic properties of scientific text (e.g., \textsc{SciNLI} and \textsc{MSciNLI}) in order to improve the performance of downstream tasks in this domain. Furthermore, we observe that intermediate training with a scientific NLI objective only using \textsc{SciNLI} fails to improve the performance of the downstream tasks. Therefore, while intermediate training with a scientific NLI objective can aid in improving the performance of downstream tasks, the diversity in the data is essential.

\section{Conclusion \& Future Directions}
We introduce a diverse scientific NLI benchmark, \textsc{MSciNLI} derived from five  scientific domains. We show that \textsc{MSciNLI} is more difficult to classify than the only other related dataset, \textsc{SciNLI}. We establish strong baselines on \textsc{MSciNLI} and find that our dataset is challenging for both PLMs and powerful LLMs. Furthermore, we provide a comprehensive investigation into the performance of scientific NLI models under domain-shift at test time and their usage in downstream NLP tasks. In the future, we will develop methods to improve the construction of prompts that enable better reasoning and inference capabilities of LLMs.

\section*{Acknowledgements}  This research is supported by NSF CAREER award 1802358 and NSF IIS award 2107518. Any opinions, findings, and conclusions expressed here are those of the authors and do not necessarily reflect the views of NSF. 
We thank our anonymous reviewers for their constructive feedback, which helped improve the quality of our paper.

\section*{Limitations}
 
From our experiments, we can see that the performance of the LLMs is low (best performing Macro F1 is $51.77\%$) on \textsc{MSciNLI}, which shows a lot of room for future improvement. The design of the prompts have a high impact on the performance as we can see from the results, thus, further exploration of other prompting strategies can potentially improve the performance further. In the future, we will focus on the design of other prompts to boost the performance of LLMs in scientific NLI.

\bibliography{anthology, custom}
\bibliographystyle{acl_natbib}

\newpage
\newpage
\clearpage

\appendix

\section{Additional Dataset Details}

\begin{table}
\centering
\small

\begin{tabular}{lp{0.3\textwidth}}
\toprule
\textbf{Class} & \textbf{Linking Phrases}\\
\midrule
\textsc{Contrasting} & ‘However’, ‘On the other hand', ‘In contrast', ‘On the contrary'\\
 \midrule
\textsc{Reasoning} & ‘Therefore', ‘Thus', ‘Consequently’, ‘As a result', ‘As a consequence', ‘From here, we can infer’\\
\midrule
\textsc{Entailment} & ‘Specifically’, ‘Precisely’, ‘In particular’, ‘Particularly’, ‘That is’, ‘In other words’\\
\bottomrule
\end{tabular}
\caption{Linking phrases used to extract sentence pairs and their corresponding classes.}
\label{table:linking_phrases}
\end{table}

\label{appendix:data_construction}

\subsection{More Details about Data Sources}
\label{appendix:data_sources}
To construct our dataset, for all five domains, we choose papers published after the year 2000. In particular, the sentence pairs for the training set of \textsc{NeurIPS} are extracted from papers published between 2000 and 2018 and the test and development sets are derived from the papers published in 2019. The training sets for the four ACM domains---\textsc{Hardware}, \textsc{Networks}, \textsc{Software \& its Engineering}, and \textsc{Security \& Privacy} are constructed from the papers published between 2000 and 2014. The sentence pairs extracted from the papers published between 2015 and 2017 are used to create the test and development sets for each domain.

\subsection{List of Linking Phrases}
\label{appendix:list_of_phrases}
To construct \textsc{MSciNLI}, we use the same list of linking phrases and their corresponding classes as \textsc{SciNLI}. Table \ref{table:linking_phrases} shows the linking phrases and their classes.

\subsection{Details about Annotators and Annotation Instructions}
\label{annotator_details}

In this section, we provide the details about the annotators we hired for constructing the test and development sets of \textsc{MSciNLI} (\S \ref{annotators_construction}), and for evaluating the human performance (\S \ref{annotator_evaluation}). 

\subsubsection{Annotators for constructing the test and development sets.}
\label{annotators_construction}
For constructing the \textsc{MSciNLI} development and test sets (in Section \ref{sec:manual_annotation}), we hired $7$ computer science undergraduate students as research interns at our institution who were compensated in an hourly basis by $\$15$/hour. Each annotator was trained with several pilot iterations before they started the final annotations for constructing the dataset. Moreover, out of the $7$ students that we initially hired, only $3$ were selected as the final annotators based on their performance during training to ensure a high quality of labels in our dataset.

The training phase of the students consists of $3$ iterations. At each iteration, all $7$ students were given a pilot batch and were instructed to predict the label based on the two sentences in each sample. We provide feedback to all students at the end of each iteration. In addition to the hired students, an author of this paper also annotated the examples in the third training iteration. $3$ students were then selected as the final annotators who have the top three agreement rates with the author ($79.3\%$, $78.6\%$, $75.8\%$). Once the annotators are trained, they start the final annotations to create the benchmark evaluation set of \textsc{MSciNLI}. 

Note that the annotators are instructed to label each pair of sentences based on the four scientific NLI relations and not based on what could be a possibly good linking phrase between them. This annotation instruction ensures that the scientific NLI task formulation remains the same as the traditional NLI task---predicting the semantic relation between a pair of sentences.

\subsubsection{Annotators for evaluating human performance.} 
\label{annotator_evaluation}

For evaluating the human performance on \textsc{MSciNLI} (in Section \ref{sec:human_performance}), we hire expert as well as non-expert annotators via a crowd-sourcing platform called COGITO.\footnote{\url{https://www.cogitotech.com/}} We ensured that none of the annotators for evaluating the human performance is involved with the construction of \textsc{MSciNLI} at any capacity. We distinguish between the expert and non-expert annotators based on whether they have the relevant background on the scientific domains in \textsc{MSciNLI}. Both sets of annotators are trained in the same fashion as the annotators who helped construct the test and development sets (described in the previous paragraphs). Both expert and non-expert annotators are paid at a rate of $\$0.6$/sample.  

\setlength\dashlinedash{0.2pt}
\setlength\dashlinegap{1.5pt}
\setlength\arrayrulewidth{0.3pt}
\begin{table}[t]
\centering
\small

  \begin{tabular}{l c c }
    \toprule
{\bf Class} & {\bf \#Annotated}     & {\bf Agreement} \\ 
   \midrule
    Contrasting & $1748$ & $92.9\%$ \\
    Reasoning & $1748$ & $83.1\%$ \\
    Entailment & $1748$ & $79.2\%$ \\
    Neutral & $1748$ & $96.7\%$ \\
    \hdashline
    Overall & $6992$ & $88.0\%$\\
    \bottomrule
  \end{tabular}

  \caption{\small Number of manually annotated examples and the agreement rate between the gold labels and automatically assigned labels for each class.
  }

    \label{table:class_wise_agreement_table}
    
\end{table}

\subsection{Class-wise Agreement Rate}
\label{appendix:class_wise_agreement}
The total number of annotated examples while constructing the test and development sets (in Section \ref{sec:manual_annotation}) for each class and the agreement rate between the gold label and the automatically assigned label based on linking phrases can be seen in Table \ref{table:class_wise_agreement_table}. 

\subsection{Difference/Closeness of the Domains} We quantify the differences/closeness of the domains in \textsc{MSciNLI} and the computational linguistic domain from \textsc{SciNLI} as the pairwise cosine similarities of the probability distributions of the RoBERTa-base\footnote{\url{https://huggingface.co/roberta-base}} vocabulary over each domain. The cosine similarities are reported in Table \ref{table:closeness}. We can see that the first four domains in the Table show a high similarity among them. Recall that the sentence pairs for all of these four domains are extracted from papers published in the ACM digital library. The high cosine similarities illustrate that the writing style and the vocabulary used in these domains are similar. In contrast, the cosine similarity of \textsc{NeurIPS} is the lowest with all other four domains in \textsc{MSciNLI}. Therefore, the vocabulary and the writing style in the papers published in \textsc{NeurIPS} differs substantially from the other four domains. Furthermore, it can be seen that the similarity between \textsc{ACL} and the five domains in \textsc{MSciNLI} is low, which illustrates that our dataset indeed diversifies the task of scientific NLI.

\section{Implementation Details}
\label{appendix:implementation_details}

All of our experiments are implemented using PyTorch.\footnote{\url{https://pytorch.org/}} The details are provided below.

\paragraph{\textsc{BiLSTM} baseline} Two separate BiLSTM layers are used to get the sentence level representations of the two sentences in each pair. The token embeddings of each sentence are sent through the respective BiLSTM layer and then the output hidden states are averaged to get the sentence level representations. 
The context vector $S_c$ is derived by concatenating the sentence level representations, their element-wise multiplication and difference.
$S_c$ is projected with a weight matrix $\mathbf{W} \in \mathbb{R}^{d \times 4}$ by using a linear layer with softmax to predict the class.

Each BiLSTM layer is equipped with 300D Glove \cite{pennington2014glove} embeddings which are allowed to be updated during training. The hidden state size for both BiLSTM layers is set at 300. The models are trained for 30 epochs with early stopping where we set the patience to be 10. The Macro F1 of the development score in every epoch is used as the stopping criteria. We use a cross-entropy loss and Adam optimizer \cite{kingma2014adam} to optimize the model parameters.
The min-batch size and learning rate are set at 64 and 0.001, respectively. 

\setlength\dashlinedash{0.2pt}
\setlength\dashlinegap{1.5pt}
\setlength\arrayrulewidth{0.3pt}
\begin{table}[t]
\centering
\small

  \begin{tabular}{l c c c c c c}
    \toprule
{\bf } & {\bf \textsc{HW}}  & {\bf \textsc{NW}} & {\bf \textsc{SWE}} & {\bf \textsc{Sec}} & {\bf \textsc{NIPS}} & {\bf \textsc{ACL}}\\ 
   \midrule
    {\bf \textsc{HW}} & $1$ \\
    {\bf \textsc{NW}} & $0.94$ & $1$ \\
    {\bf \textsc{SWE}} & $0.95$ & $0.95$ & $1$\\
    {\bf \textsc{Sec}} & $0.93$ & $0.96$ & $0.97$ & $1$ \\
    {\bf \textsc{NIPS}} & $0.70$ & $0.63$ & $0.63$ & $0.61$ & $1$\\
    {\bf \textsc{ACL}} & $0.76$ & $0.69$ & $0.71$ & $0.68$ & $0.81$ & $1$\\
    \bottomrule
  \end{tabular}

  \caption{\small Pair-wise cosine similarities of the probability distributions of the vocabulary of RoBERTa-base over domain-wise training sets. Here, \textsc{HW:Hardware}, \textsc{NW:Networks}, \textsc{SWE: Software \& its Engineering.}, \textsc{Sec: Security \& Privacy},  \textsc{NIPS: NeurIPS}, and \textsc{ACL:} data from \textsc{SciNLI.}}
    \vspace{-2mm}
    \label{table:closeness}
    
\end{table}
\paragraph{PLM baselines} The details of our pre-trained models are described as follows: (a) \textbf{\textsc{BERT}} \cite{devlin-etal-2019-bert} - pre-trained by masked language modeling (MLM) and Next Sentence Prediction (NSP) objectives on BookCorpus \cite{zhu2015aligning} and Wikipedia; (b)  \textbf{\textsc{SciBERT}} \cite{beltagy-etal-2019-scibert} - pre-trained using the same objectives as BERT but using scientific text exclusively as the pre-training data; (c) \textbf{\textsc{RoBERTa}} \cite{liu2019roberta} - an extension of BERT which uses a variation of MLM where different words are masked in each epoch dynamically (unlike static masking in standard MLM). It is also trained on larger amount of text, larger mini-batch size and larger number of epochs compared to BERT; and (d) \textbf{\textsc{XLNet}} \cite{yang2019xlnet} - pre-trained with a ``Permutation Language Modeling'' objective instead of MLM to provide bi-directional context to the model while being auto-regressive.

\begin{table*}[t]
\centering
\small

\scalebox{1.00}{
  \begin{tabular}{l c c c c c c}
    \toprule

&  {\textsc{Hardware}} &  {\textsc{Networks}} & {\textsc{SWE}} & {\textsc{Security}} & {\textsc{NeurIPS}} & {\textsc{Overall}}\\

    \midrule
    \textbf{\textsc{Domain-wise}}\\

    \mbox{{\color{white}{x}}}\textsc{BERT} & $68.63 \pm 1.4$ & $67.69 \pm 1.3$ & $65.75 \pm 0.8$ & $67.04 \pm 2.9$ & $70.48 \pm 1.6$ & -\\
    \mbox{{\color{white}{x}}}\textsc{SciBERT} & $72.76 \pm 0.9$ & $72.97 \pm 1.4$ & $72.43 \pm 0.8$ & $72.45 \pm 1.4$ & $76.14 \pm 0.5$ & -\\
    \mbox{{\color{white}{x}}}\textsc{XLNet} & $72.88 \pm 0.6$ & $68.85 \pm 2.2$ & $71.52 \pm 2.1$ & $71.60 \pm 0.8$ & $72.96 \pm 0.8$ & -\\
    \mbox{{\color{white}{x}}}\textsc{RoBERTa} & $74.93 \pm 1.4$ & $73.31 \pm 1.7$ & $74.44^{*} \pm 0.5$ & $74.45 \pm 2.2$ & $76.02^{\#} \pm 1.1$ & -\\
    \midrule
    
    \textbf{\textsc{Merged - small}}\\

    \mbox{{\color{white}{x}}}\textsc{BERT} & $69.36 \pm 0.5$ & $68.08 \pm 1.3$ & $66.61 \pm 0.1$ & $67.66 \pm 1.4$ & $71.62 \pm 0.1$ & $68.67 \pm 0.3$\\
    \mbox{{\color{white}{x}}}\textsc{SciBERT} & $72.95 \pm 0.3$ & $72.88 \pm 1.1$ & $72.66 \pm 0.2$ & $72.37 \pm 1.5$ & $74.96 \pm 1.6$ & $73.17 \pm 0.7$\\
    \mbox{{\color{white}{x}}}\textsc{XLNet} & $72.87 \pm 1.7$ & $71.03 \pm 1.8$ & $72.21 \pm 1.7$ & $70.90 \pm 0.8$ & $73.45 \pm 0.7$ & $71.96 \pm 1.2$\\
    \mbox{{\color{white}{x}}}\textsc{RoBERTa} & $75.06^{*} \pm 0.7$ & $73.20 \pm 1.1$ & $74.49^{*} \pm 0.4$ & $73.73^{\#} \pm 1.1$ & $75.75 \pm 1.6$ & $74.47 \pm 0.8$\\
    \midrule
    
    \textbf{\textsc{Merged - large}}\\

    \mbox{{\color{white}{x}}}\textsc{BERT} & $72.89 \pm 0.1$ & $74.10 \pm 1.3$ & $71.37 \pm 0.3$ & $72.38 \pm 2.5$ & $75.46 \pm 0.8$ & $73.24 \pm 0.8$\\
    \mbox{{\color{white}{x}}}\textsc{SciBERT} & $75.91 \pm 0.1$ & $76.51 \pm 0.5$ & $75.28 \pm 1.1$ & $75.94 \pm 0.4$ & $78.78 \pm 0.1$ & $76.48 \pm 0.4$\\
    \mbox{{\color{white}{x}}}\textsc{XLNet} & $75.59 \pm 0.5$ & $75.25 \pm 0.1$ & $73.98 \pm 0.6$ & $75.09 \pm 0.8$ & $77.64 \pm 1.0$ & $75.51 \pm 0.3$\\
    \mbox{{\color{white}{x}}}\textsc{RoBERTa} & $77.79^{\$} \pm 0.2$ & $75.45 \pm 1.5$ & $77.10^{\#} \pm 0.7$ & $77.71^{\$} \pm 0.2$ & $78.04 \pm 0.8$ & $77.21^{\#} \pm 0.3$\\
    \bottomrule
  \end{tabular}}

   \caption{\small Macro F1 scores (\%) of our PLM baselines on different domains trained in different settings. Here, \textsc{SWE}: Software \& its Engineering and \textsc{Security}: Security \& Privacy. All \textsc{Merged - small} scores are statistically indistinguishable from their \textsc{Domain-wise} counterparts according to a paired t-test with $p < 0.05$. All \textsc{Merged-large} scores show statistically significant improvement over \textsc{Merged-small}.
  $^*$, $^\#$, $^\$$ indicate statistically significant improvement by \textsc{RoBERTa} over \textsc{SciBERT}, \textsc{XLNet}, and both \textsc{SciBERT} and \textsc{XLNet}, respectively. }
  
    \label{table:domain_wise_vs_merged}
\end{table*}

For these PLM baselines, the two sentences in each example are concatenated with a \texttt{[SEP]} token between them to be used as the input and the hidden representation embedded in the \texttt{[CLS]} token is then projected with a weight matrix $\mathbf{W} \in \mathbb{R}^{d \times 4}$. Finally, we use softmax on the projected representation to get the probability distribution over the four classes. The class with the maximum probability is predicted as the label for each input pair.

Each PLM baseline is fine-tuned for 10 epochs with early stopping using the huggingface\footnote{\url{https://huggingface.co}} library. The patience for early stopping is set at 2. The learning rate and the mini-batch size is set at $2e-5$, and $64$, respectively. We use a cross-entropy loss and Adam optimizer \cite{kingma2014adam} to optimize the model parameters.

\paragraph{LLM baselines} We make use of the prompt templates described in Section \ref{sec:LLM_baselines} to construct the inputs to the LLM baselines. Similar to the PLM baselines, we conduct our experiments for LLM baselines using the huggingface library. We employ a greedy decoding strategy with a maximum generated token count to be $40$. Generally, instead of only providing the answer to our multiple-choice question, the LLMs generates a more verbose response with the answer contained in it. We manually examine the responses for each prompt by each LLM and develop scripts to extract the correct answer with rule-based approaches. 

\paragraph{Computational Cost} 
The BiLSTM and PLM experiments are conducted on a single NVIDIA RTX A5000 GPU. The BiLSTM model was trained in $\approx 30$ minutes. The time needed to fine-tune each PLM baseline on the full \textsc{MSciNLI} training set using a single GPU is $\approx 2$ hours. The inference by the LLM baselines is conducted using two  NVIDIA RTX A5000 GPUs and it took  $\approx 3$ hours on average for each experiment.

\section{Additional Results}
\label{appendix:additional_results}

\subsection{Domain-wise vs Merged}  In addition to fine-tuning on the combined \textsc{MSciNLI} training set ($127,320$ examples) in Section \ref{sec:PLM_baselines}, we experiment with the PLM baselines in two other settings: \textsc{Domain-wise} and \textsc{Merged-small} (see the description of these settings below) and compare their performance with the model fine-tuned on the combined \textsc{MSciNLI} training set denoted as \textsc{Merged-large}. The motivation behind these experiments is two-fold: a) to understand the impact of diversity of examples in model training (\textsc{Domain-wise} vs \textsc{Merged-small}) (when the models are trained on data from a single domain vs data from diverse domains---but all being trained on the same training set size); and b) to understand the impact of training set size (\textsc{Merged-small} vs \textsc{Merged-large}). In the \textsc{Domain-wise} setting, we train and evaluate separate models for each domain using the data from the respective domain. For the \textsc{Merged-small} setting, we randomly down-sample the training set of each domain to $5092$ examples (class-balanced) before combining them to ensure that the total size of the merged set \textsc{Merged-small} is similar to the \textsc{Domain-wise} training set size ($\approx 25,464$). We combine the downsampled data from all domains and train {\em a single model} using the merged data. This model is then evaluated on the test set of each domain and the combined \textsc{MSciNLI} test set. The \textsc{Merged-large} setting corresponds to the combined training set of MSciNLI of $127,320$ examples. 

We run each experiment three times and report the average and standard deviation of the Macro F1 score of the models in the three settings in Table \ref{table:domain_wise_vs_merged}. We find the following:

\paragraph{Training models on diverse data is more optimal.}
We can see that each model trained in the \textsc{Merged - small} setting shows similar performance as their \textsc{Domain-wise} counterparts. Moreover, in some cases (e.g., for \textsc{BERT} for all domains, \textsc{XLNet} for \textsc{NeurIPS}), the \textsc{Merged - small} models outperform the \textsc{Domain-wise} models. Recall that the size of the \textsc{Merged-small} training set is the same as each of the \textsc{Domain-wise} training sets. Since we train separate models for each domain in the \textsc{Domain-wise} setting, it is five times computationally more expensive than the \textsc{Merged - small} setting where a single model is trained for all domains. Therefore, training a single model on diverse data can reduce the computational cost without compromising model performance resulting in a more optimal approach.

\paragraph{More data leads to better performance.} 
Next we compare the performance of the model fine-tuned on \textsc{Merged-small} with its \textsc{Merged-large} counterpart. The results show that \textsc{Merged - large} models consistently outperform the \textsc{Merged - small} models by a substantial margin. Therefore, the performance on our dataset improves with the the increase of dataset size.

\begin{table*}[t]
\centering
\small

\scalebox{1.00}{
  \begin{tabular}{l c c c c c}
    \toprule

{\textsc{Model}} &  {\textsc{Contrasting}} &  {\textsc{Reasoning}} & {\textsc{Entailment}} & {\textsc{Neutral}} & {\textsc{Macro Ave.}}\\

    \midrule
    \textbf{Precision}\\
    \mbox{{\color{white}{x}}}\textsc{RoBERTa} & $74.24$ & $74.61$ & $76.38$ & $85.46$ & $77.67$\\
    \mbox{{\color{white}{x}}}\textsc{Llama-2} & $56.35$ & $36.19$ & $49.76$ & $71.96$ & $53.57$\\
    \hdashline
    \textbf{Recall}\\
    \mbox{{\color{white}{x}}}\textsc{RoBERTa} & $87.93$ & $67.36$ & $74.63$ & $79.36$ & $77.32$\\
    \mbox{{\color{white}{x}}}\textsc{Llama-2} & $54.50$ & $50.10$ & $42.30$ & $57.50$ & $51.10$\\
    \hdashline
    \textbf{F1}\\
    \mbox{{\color{white}{x}}}\textsc{RoBERTa} & $80.43$ & $70.76$ & $75.44$ & $82.22$ & $77.21$\\
    \mbox{{\color{white}{x}}}\textsc{Llama-2} & $55.41$ & $42.03$ & $45.72$ & $63.92$ & $51.77$\\
    \bottomrule
    
  \end{tabular}}

   \caption{\small Class-wise precision (\%), recall (\%), F1 (\%), and their macro averages (\%) of our best performing PLM and LLM baselines on \textsc{MSciNLI}.
  }
  
    \label{table:class_wise_results}
\end{table*}

\begin{table*}[t]
\centering
\small

\scalebox{1.00}{
  \begin{tabular}{l c c c c c c}
    \toprule
&  {\textsc{Hardware}} &  {\textsc{Networks}} & {\textsc{SWE}} & {\textsc{Security}} & {\textsc{NeurIPS}} & {\textsc{Overall}}\\

    \midrule
    
    \textbf{\textsc{GPT-NeoXT-Chat}}\\
    \mbox{{\color{white}{x}}}\textsc{Prompt - 1$_{zs}$} & $17.84$ & $19.17$ & $20.19$ & $18.04$ & $16.99$ & $18.49$\\
    \mbox{{\color{white}{x}}}\textsc{Prompt - 2$_{zs}$} & $20.48$ & $18.28$ & $20.67$ & $21.62$ & $27.56$ & $22.14$\\
    \mbox{{\color{white}{x}}}\textsc{Prompt - 3$_{zs}$} & $12.69$ & $15.30$ & $13.63$ & $13.90$ & $14.66$ & $14.12$\\
    \hdashline
    \mbox{{\color{white}{x}}}\textsc{Prompt - 1$_{fs}$} & $10.00$ & $10.00$ & $10.00$ & $10.00$ & $10.00$ & $10.00$\\
    \mbox{{\color{white}{x}}}\textsc{Prompt - 2$_{fs}$} & $10.00$ & $10.00$ & $10.00$ & $10.00$ & $10.00$ & $10.00$\\
    \mbox{{\color{white}{x}}}\textsc{Prompt - 3$_{fs}$}  & $10.00$ & $10.00$ & $10.00$ & $10.00$ & $10.00$ & $10.00$\\

    \bottomrule
    
  \end{tabular}}

   \caption{\small Macro F1 scores (\%) of our \textsc{GPT-NeoX} baseline on different domains. Here, \textsc{SWE}: Software \& its Engineering and \textsc{Security}: Security \& Privacy. 
  }
  
    \label{table:baseline_results_gpt_neox}
\end{table*}

\subsection{Class-wise Performances}
We evaluate the class-wise performance of our best performing PLM baseline---\textsc{RoBERTa} trained on the combined \textsc{MSciNLI} training set, and our best performing LLM baseline---\textsc{Llama-2} with \textsc{Prompt - 3} in the few-shot setting. The results are reported in Table \ref{table:class_wise_results}. 

As we can see, both models show a better performance for the \textsc{Contrasting} and the \textsc{Neutral} classes, and they struggle more for the \textsc{Reasoning}, and \textsc{Entailment} classes. However, even the \textsc{Contrasting} and the \textsc{Neutral} classes are still challenging for the models with substantial headroom for improvement.

\subsection{Another LLM Baseline - \textsc{GPT-NeoX}} In addition to the LLMs that we explore in Section \ref{sec:baselines}, we also experiment with the \textsc{GPT-NeoXT-Chat-Base-20B}\footnote{\url{https://huggingface.co/togethercomputer/GPT-NeoXT-Chat-Base-20B}} variant of the \textsc{GPT-NeoX} model. However, despite being much larger in size than the \textsc{Llama-2} and \textsc{Mistral} baselines (20 billion parameters vs 13B and 7B, respectively), \textsc{GPT-NeoX} failed to show any promising performance (for the same three prompts used in the paper). We report the performance of these baselines in Table \ref{table:baseline_results_gpt_neox}. We can see that the best performance for \textsc{GPT-NeoX} is shown by \textsc{\textsc{Prompt - 1$_{zs}$}} with an overall Macro F1 of only $22.14\%$. Moreover, none of the few-shot versions of the prompts shows any meaningful performance for this model ($10.00\%$ in Macro F1 with four labels in total means that the model always predicts the same label). In our future work, we will focus on the designing of other prompts that can improve the performance of the LLMs.  

\setlength\dashlinedash{0.2pt}
\setlength\dashlinegap{1.5pt}
\setlength\arrayrulewidth{0.3pt}
\begin{table}[t]
\centering
\small

  \begin{tabular}{ r r c c }
    \toprule

& &  \multicolumn{2}{c}{\bf \textsc{SciNLI}} \\  \cmidrule(lr){3-4} 
   {\bf Model} & & {\bf F1}     & {\bf Acc} \\ 
   \midrule

   RoBERTa & \mbox{{\color{white}{xxx}}}\textsc{both sentences} & $77.21$ & $77.20$ \\
   & \mbox{{\color{white}{xxx}}}\textsc{only $2^{nd}$ sentence} & $52.55$ & $53.55$ \\
   \midrule
   SciBERT &  \mbox{{\color{white}{xxx}}}\textsc{both sentences} & $76.48$ & $76.46$ \\
   & \mbox{{\color{white}{xxx}}}\textsc{only $2^{nd}$ sentence} & $53.14$ & $53.65$ \\
    \bottomrule
  \end{tabular}

  \caption{\small Performance comparison on \textsc{MSciNLI} when both sentences are concatenated vs. when only second sentence is used as the input.}
\vspace{-3mm}
    \label{table:artifacts_balanced}
\end{table}

\subsection{Only-Second-Sentence Baseline}
\label{appendix:only_second}
To evaluate the degree of spurious correlations \cite{gururangan-etal-2018-annotation} that may exist in \textsc{MsciNLI}, we experiment with \textit{only-second-sentence} models. Specifically, we fine-tune both \textsc{RoBERTa} and \textsc{SciBERT} where only the second sentence is used as the input. A comparison between the \textit{only-second-sentence} models and the models using both sentences can be seen in Table \ref{table:artifacts_balanced}. The results show that the performance decreases by a large margin when only the second sentence is used as the input. Therefore, the amount of spurious correlation in \textsc{MSciNLI} is smaller compared with other existing NLI datasets (e.g., SNLI \cite{bowman-etal-2015-large}) and the models need to learn the semantic relation between the sentences in each pair in order to perform well.

However, given that the performance of the \textit{only-second-sentence} models are much higher than chance ($25$\%), we believe there are still some degree of spurious patterns in \textsc{MSciNLI}. In our future work, we will explore methods to identify and reduce the degree of spurious patterns in scientific NLI.

\setlength\dashlinedash{0.2pt}
\setlength\dashlinegap{1.5pt}
\setlength\arrayrulewidth{0.3pt}
\begin{table*}[t]
\centering
\small

  \begin{tabular}{l p{40em}}
    \toprule
    {\bf \textsc{Prompt - 1}} & <human>: Consider the following two sentences:\newline
Sentence1: <sentence1>\newline
Sentence2: <sentence2>\newline
What is the semantic relation between Sentence1 and Sentence2? Choose from the following options: 1. Entailment, 2. Reasoning, 3. Contrasting, 4. Neutral.\newline
<bot>: \\
\midrule
    {\bf \textsc{Prompt - 2}} & <human>: Consider the following class definitions of four semantic relations between a pair of sentences.\newline
Entailment: <definition of entailment>\newline
Contrasting: <definition of contrasting>\newline
Reasoning: <definition of reasoning>\newline
Neutral: <definition of neutral>\newline
\newline
Now consider the following two sentences: \newline
Sentence1: <sentence1>\newline
Sentence2: <sentence2>\newline
Based on only the information available in these two sentences and the class definitions, answer the following:
What is the semantic relation between Sentence1 and Sentence2? Choose from the following options: 1. Entailment, 2. Reasoning, 3. Contrasting, 4. Neutral.\newline
<bot>: \\

\midrule
    {\bf \textsc{Prompt - 3}} & <human>: Consider the following two sentences:\newline
    Sentence1: <sentence1>\newline
Sentence2: <sentence2>\newline
Based on only the information available in these two sentences, which of the following options is true?\newline
a. Sentence1 generalizes, specifies or has an equivalent meaning with Sentence2.\newline
b. Sentence1 presents the reason, cause, or condition for the result or conclusion made Sentence2.\newline
c. Sentence2 mentions a comparison, criticism, juxtaposition, or a limitation of something said in Sentence1.\newline
d. Sentence1 and Sentence2 are independent.\newline
<bot>:\\
    \bottomrule
  \end{tabular}

  \caption{\small Prompt templates used for our experiments with LLMs. Here, <X> indicates a placeholder X which is replaced in the actual prompts. }

    \label{table:prompts}
    
\end{table*}

\section{Prompts for LLMs}
\label{appendix:prompts}
The zero-shot versions of the three prompt templates that we construct for LLMs can be seen in Table \ref{table:prompts}.  For the few-shot versions of the prompts, we pre-pend four input-human annotated output exemplars (one for each class) to each prompt. Note that the \textit{<human>} and \textit{<bot>} tags in the prompts in the Table are replaced with the relevant tags for each LLM (e.g., \texttt{[INST]}).

\begin{figure*}[t]
\centering
    \includegraphics[scale=0.4]{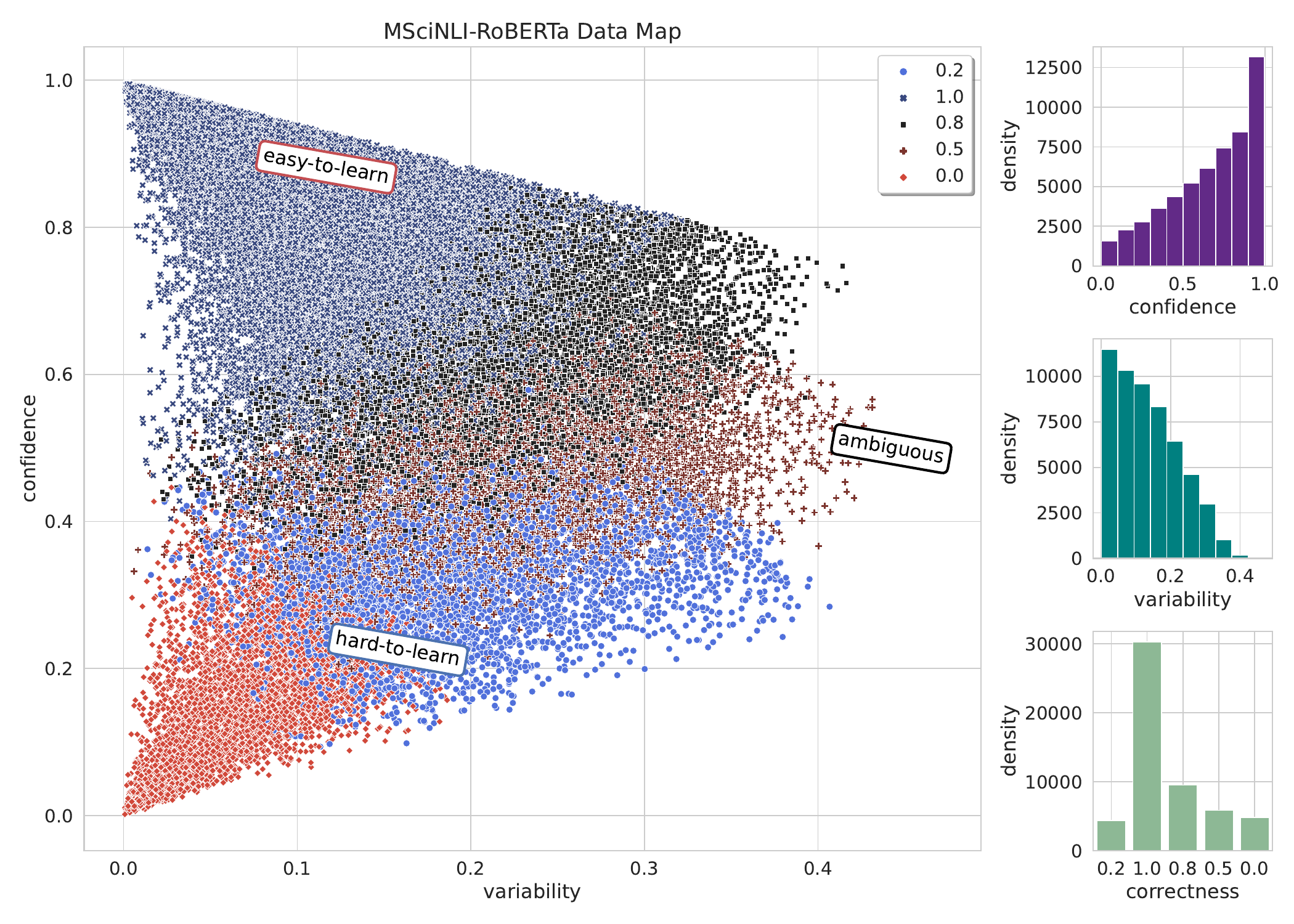}

    \caption{\small Data cartography of \textsc{MSciNLI}. The colors and shapes indicate the correctness of each example.}

    \label{fig:MSciNLI_data_map}
\end{figure*}

\section{Training Dynamics Based Data Selection}
\label{appendix: training_dynamics}
The \textit{easy/hard/ambiguous} subsets of the training data are selected based on their training dynamics \cite{swayamdipta-etal-2020-dataset}. Specifically, the training dynamics of each example is defined in the form of three metrics---\textit{confidence}, \textit{variability}, and \textit{correctness} during training a classifier. These metrics are used to plot the examples in a data map to perform a visual analysis. The aforementioned three subsets of the training set are then selected based on \textit{confidence} and \textit{variability}. In this section, we define these metrics, perform a data cartography of \textsc{MSciNLI}, and describe the method to select the subsets used in Section \ref{sec:cartography}.

\subsection{Metrics Definitions} 
The \textit{confidence} of each example is defined as the average of the probability predicted by a classifier for its label over the training epochs. That is, for a training example $X_i$ and its label $y_i$, the confidence $c_i$ is calculated as follows: 
\begin{equation}
\label{eqn:confidence_calculation}
    c_{i} = \frac{1}{E}\sum_{e=1}^{E}p(y_{i} | X_{i}, \theta^{e})
\end{equation}
Here, $E$ is the number of training epochs, $\theta_{e}$ is the model at epoch $e$ and $p$ is the probability of the label given $X_i$ and $\theta_{e}$. The \textit{variability} of each example is defined as the standard deviation of the predicted probability for its label over the training epochs. More formally, the variability, $v_i$ of an example $X_i$ is calculated as: 
\begin{equation}
\label{eqn:variability_calculation}
    v_{i} = \sqrt{\frac{\sum^{E}_{e=1}(p(y_{i} | X_{i}, \theta^{e})- c_{i})^{2}}{E}}
\end{equation}
Finally, the fraction of the training epochs where the classifier predicts the label of an example correctly is defined as its \textit{correctness}.

\subsection{Data Plot}
\vspace{-2mm}
For creating the data plot, we fine-tune a \textsc{RoBERTa} classifier on the combined \textsc{MSciNLI} training set. While training, we record the probability distributions predicted by the classifier for the training examples over the four labels in each epoch. We then calculate the confidence, variability, and correctness of each example using the recorded probability distributions and plot them in the data map based on these calculated values. The data plot can be seen in Figure \ref{fig:MSciNLI_data_map}.

We can see that the model shows a high \textit{correctness} for the examples in the high \textit{confidence} region. Therefore, the examples in this region are \textit{easy-to-learn} for the model. On the other hand, the plot shows that the \textit{correctness} of the model's predictions is very low in the low \textit{confidence} region of the map. Thus, the examples in this region are \textit{hard-to-learn} for the model. Since, by definition, the probability predicted by the model shows a high fluctuation for the examples in the high \textit{variability} region, they can be denoted as \textit{ambiguous} examples. 

Based on these observations from the data map, we select the various subsets from the full training set as follows.

\subsection{Data Subset Selection} 

We rank the full training set based on \textit{confidence} in a \textit{descending} order and then select the top $33\%$ examples as the $33\%$ \textit{easy-to-learn} subset. Similarly, we rank the full training set based on \textit{confidence} in an \textit{ascending} order and then select the top $33\%$ examples as the $33\%$ \textit{hard-to-learn} subset. The top $33\%$ examples from a ranking based on the \textit{variability} in a \textit{descending} order is chosen as the top $33\%$ \textit{ambiguous} subset. For the `$100\% -$ top $25\%$ \textit{hard}' and `$100\% -$ top $5\%$ \textit{hard}' subsets, we remove top $5\%$ and $25\%$ examples from the ranking based on \textit{confidence} in an \textit{ascending} order, respectively from the full training set.

\begin{table*}[t]
\centering
\small

\scalebox{1.00}{
  \begin{tabular}{l c c c c c c}
    \toprule

{{\bf \backslashbox{Train}{Test}}} &  {\textsc{Hardware}} &  {\textsc{Networks}} & {\textsc{SWE}} & {\textsc{Security}} & {\textsc{NeurIPS}} & {\textsc{ACL}}\\

    \midrule

    \textsc{SciNLI} & $75.60 \pm 0.8$ & $72.71 \pm 0.5$ & $74.36 \pm 0.3$ & $75.00 \pm 0.3$ & $78.36 \pm 1.0$ & $78.08 \pm 0.4$\\
    \textsc{MSciNLI} & $77.79 \pm 0.2$ & $75.45 \pm 1.5$ & $\textbf{77.10} \pm \textbf{0.7}$ & $77.71 \pm 0.2$ & $78.04 \pm 0.8$ & $76.74 \pm 0.5$\\
    \textsc{MSciNLI+} & $\textbf{77.99} \pm \textbf{0.4}$ & $\textbf{77.48} \pm \textbf{0.4}$ & $76.78 \pm 1.1$ & $\textbf{78.08} \pm \textbf{1.4}$ & $\textbf{80.02} \pm \textbf{1.4}$ & $\textbf{79.48} \pm \textbf{0.4}$\\

    \midrule

  \end{tabular}}

   \caption{\small Macro F1 scores (\%) of the cross-dataset models based on \textsc{RoBERTa} on different domains. Here, \textsc{SWE}: Software \& its Engineering and \textsc{Security}: Security \& Privacy. Best scores are in bold. 
  }
  
    \label{table:cross_dataset_domainwise}
\end{table*}

\setlength\dashlinedash{0.2pt}
\setlength\dashlinegap{1.5pt}
\setlength\arrayrulewidth{0.3pt}
\begin{table}[t]
\centering
\small
  \begin{tabular}{l c c c c}
    \toprule
    {\bf \backslashbox{Tr}{Te}} & {\bf SciNLI} & {\bf MSciNLI} & {\bf SciTail} & {\bf MNLI}\\
   \toprule
    \textsc{SciNLI} & $86.03$ & $81.43$ & $51.62$ & $53.63$\\
     \textsc{MSciNLI} & $83.18$ & $82.56$ & $55.66$ & $58.72$\\
     \textsc{SciTail} & $48.64$ & $48.86$ & $91.19$ & $73.42$\\
    \textsc{MNLI} & $45.40$ & $47.57$ & $78.18$ & $91.31$\\

    \bottomrule
  \end{tabular}
  \caption{\small Cross dataset performances (Macro F1 ($\%$)) of \textsc{RoBERTa} on different datasets in a 2-class setting.}
    \label{table:gen_domain_vs_scientific_domain}

\end{table}

\section{Additional Cross-dataset Experiments}
\label{sec:cross_dataset_additional}
\subsection{Domain-wise Performance by Cross-dataset Models}
\label{sec:cross_dataset_domainwise}
We can see the domain-wise performance of the cross-dataset models described in Section \ref{sec:cross_dataset} in Table \ref{table:cross_dataset_domainwise}. The results show that 
\textsc{MSciNLI+} shows a better performance on domain-level as well. 

\subsection{Out-of-dataset Performance on Regular NLI datasets}
To understand the effect of data diversity in the performance of scientific NLI models on regular NLI datasets, we perform a set of experiments with RoBERTa where we train models on SciNLI, MSciNLI, SciTail \cite{khot2018scitail}, and MNLI \cite{williams-etal-2018-broad} and evaluate them on each of the test sets of these datasets. Note that since the test set of \textsc{MNLI} is not publicly available, we use the development set as the test set and a randomly sampled set of size $10,000$ as the development set. Given that the NLI classes differ in these datasets, we convert \textsc{SciNLI}, \textsc{MSciNLI} and \textsc{MNLI} into 2-class datasets. Specifically, we update the labels of all non-entailment classes, i.e., contradiction and neutral for \textsc{MNLI} and contrasting, reasoning, neutral for \textsc{SciNLI} and \textsc{MSciNLI} to a class named \textsc{Not-entailment}. We do not change any labels in \textsc{SciTail} because it is already in a 2-class setting using \textsc{Entailment} and \textsc{Not-entailment} as the classes. The Macro F1 from these experiments are in Table \ref{table:gen_domain_vs_scientific_domain}. 

We can see that the model trained on \textsc{MSciNLI} shows a substantially higher performance on both \textsc{MNLI}, and \textsc{SciTail} compared to the model trained with \textsc{SciNLI}. Therefore, training the models on diverse examples improves their reasoning capabilities which results in a better performance even for traditional NLI datasets. In our future work, we will investigate how the models trained on scientific NLI datasets behave when they are tested on \textit{easy}, \textit{ambiguous} and \textit{hard-to-learn} examples of the traditional NLI datasets.

\begin{table*}
\small

\begin{tabular}{lp{0.8\textwidth}}
\toprule
\textbf{Dataset} & \textbf{Classes}\\
\midrule
\textsc{SciHTC} & ‘General and reference’, ‘Hardware', ‘Computer systems organization', ‘Networks', ‘Software and its engineering’, ‘Theory of computation', ‘Mathematics of computing', ‘Information systems', ‘Security and privacy', ‘Human-centered computing', ‘Computing methodologies', ‘Applied computing', ‘Social and professional topics'\\
 \midrule

\textsc{Paper Field} &  `Geography', `Politics', `Economics', `Business', `Sociology', `Medicine', `Psychology'\\
\midrule
\textsc{ACL-Arc} & ‘Background’, ‘Extends’, ‘Uses’, ‘Motivation’, ‘Compare/Contrast’, ‘Future work’\\
\bottomrule
\end{tabular}
\caption{\small Downstream task datasets and their classes.}
\label{table:intermideate_classes}
\end{table*}

\setlength\dashlinedash{0.2pt}
\setlength\dashlinegap{1.5pt}
\setlength\arrayrulewidth{0.3pt}
\begin{table}[t]
\centering
\small

  \begin{tabular}{l c c c}
    \toprule
{\bf Dataset} & {\bf \#Train}  & {\bf \#Test} & {\bf \#Dev} \\ 
   \midrule
    \textsc{SciHTC} & $148,928$ & $18,616$ & $18,616$ \\
    \textsc{Paper Field} & $84,000$ & $22,399$ & $5,599$ \\
    \textsc{ACL-Arc} & $1,688$ & $139$ & $114$ \\
    \bottomrule
  \end{tabular}

  \caption{ \small Number of examples in downstream tasks.
  }

    \label{table:intermideate_num_examples}
    
\end{table}

\section{Details on Intermediate Task Transfer}
\label{appendix:intermediate_transfer}
\subsection{Downstream Tasks - Dataset Details}
The categories/class labels and the number of examples in each dataset for the downstream tasks in our intermediate task transfer experiments can be seen in Tables \ref{table:intermideate_classes} and \ref{table:intermideate_num_examples}, respectively.

The details of each the downstream tasks that we experiment with are as follows.

\paragraph{SciHTC \cite{sadat-caragea-2022-scihtc}} A hierarchical multi-label scientific topic classification dataset containing $186$K papers. While each paper in \textsc{SciHTC} is assigned multiple labels from different levels of the hierarchy tree), we only consider the level 1 flat categories which are $13$ in total (see Table \ref{table:intermideate_classes}) and train the model in a multi-class (single label for each paper) setting.

\paragraph{Paper Field \cite{beltagy-etal-2019-scibert}} A paper classification dataset containing $112$K papers where each paper is classified to different scientific fields. The total number of paper classes in this dataset is $7$ (see Table \ref{table:intermideate_classes}). 

\paragraph{ACL-ARC \cite{jurgens2018measuring}} A citation intent classification dataset where the intent behind a citation made in a sentence in a scientific paper needs to be predicted. The $6$ classes in this dataset can be seen in Table \ref{table:intermideate_classes}.

\subsection{Experimental Details of Intermediate Task Transfer Learning}
In the intermediate task transfer setting, the \textsc{RoBERTa} model is trained on the NLI datasets for a single epoch (unlike the baselines). For the unsupervised intermediate training with MLM, $15\%$ tokens are randomly masked and the model is also trained for a single epoch. During the fine-tuning step, only the RoBERTa layer is initialized from the model from the intermediate training step. The parameters for the output linear layer with softmax activation is randomly initialized. The model is then fine-tuned for the downstream tasks for multiple epochs. Specifically, the models for \textsc{SciHTC}, and \textsc{Paper-field} are trained for $10$ epochs. The models for \textsc{ACL-Arc} are fine-tuned for a maximum of $20$ epochs due to its small size. Similar to our baselines, we employ early stopping with patience 2 and Macro F1 score of the development set as the stopping criteria.